\documentclass[10pt,twocolumn,letterpaper]{article}

\usepackage{cvpr}              % To produce the CAMERA-READY version
\usepackage{algorithm}
\usepackage{algpseudocode}
\usepackage{multirow}
\usepackage{makecell}
\usepackage{amsmath}
\usepackage{amssymb}
\usepackage{graphicx}
\usepackage{booktabs}
\usepackage{siunitx}
\usepackage{caption}
\usepackage{subcaption}
\usepackage{placeins}
\definecolor{cvprblue}{rgb}{0.21,0.49,0.74}
\usepackage[pagebackref,breaklinks,colorlinks,allcolors=cvprblue]{hyperref}

\def\paperID{30784}
\def\confName{CVPR}
\def\confYear{2026}

\title{How to Achieve Prototypical Birth and Death for OOD Detection?}

\author{
Ningkang Peng$^{1}$, Qianfeng Yu$^{1}$, Xiaoqian Peng$^{2}$, Linjing Qian$^{1}$, Yafei Liu$^{1}$,\\
Canran Xiao$^{3*}$, Xinyu Lu$^{1*}$, Tingyu Lu$^{4*}$, Zhichao Zheng$^{1*}$, Yanhui Gu$^{1*}$\thanks{Corresponding authors: Canran Xiao, Xinyu Lu, Tingyu Lu, Zhichao Zheng, and Yanhui Gu.}\\
$^{1}$School of Computer and Electronic Information, Nanjing Normal University\\
$^{2}$School of Artificial Intelligence and Information Technology, Nanjing University of Chinese Medicine\\
$^{3}$School of Cyber Science and Technology, Shenzhen Campus of Sun Yat-sen University, Shenzhen\\
$^{4}$Tohoku University\\
{\tt\small \{nkpeng, qfyu, qianlj, yafeiliu\}@nnu.edu.cn}\\
{\tt\small 202411148@njucm.edu.cn, xiaocr3@mail.sysu.edu.cn}\\
{\tt\small \{xy\_lu, zhengzhichao, gu\}@njnu.edu.cn}\\
{\tt\small tingyu.lu.e7@tohoku.ac.jp}
}

\begin{document}
\maketitle
\begin{abstract}
Out-of-Distribution (OOD) detection is crucial for the secure deployment of machine learning models, and prototype-based learning methods are among the mainstream strategies for achieving OOD detection. Existing prototype-based learning methods generally rely on a fixed number of prototypes. This static assumption fails to adapt to the inherent complexity differences across various categories. Currently, there is still a lack of a mechanism that can adaptively adjust the number of prototypes based on data complexity.
Inspired by the processes of cell birth and death in biology, we propose a novel method named PID (Prototype bIrth and Death) to adaptively adjust the prototype count based on data complexity. This method relies on two dynamic mechanisms during the training process: prototype birth and prototype death.
The birth mechanism instantiates new prototypes in data regions with insufficient representation by identifying the overload level of existing prototypes, thereby meticulously capturing intra-class substructures. Conversely, the death mechanism reinforces the decision boundary by pruning prototypes with ambiguous class boundaries through evaluating their discriminability.
Through birth and death, the number of prototypes can be dynamically adjusted according to the data complexity, leading to the learning of more compact and better-separated In-Distribution (ID) embeddings, which significantly enhances the capability to detect OOD samples. Experiments demonstrate that our dynamic method, PID, significantly outperforms existing methods on benchmarks such as CIFAR-100, achieving State-of-the-Art (SOTA) performance, especially on the FPR95 metric.
\end{abstract}
\section{Introduction}
\label{sec:introduction}

The safety and reliability of machine learning models, particularly deep neural networks, are considered paramount in open-world environments\cite{parmar2023open}. A core challenge is presented by the inevitable encounter with Out-of-Distribution (OOD) data, samples unseen during the training phase, when models are deployed\cite{liu2021towards}. The OOD detection task is aimed at equipping models with the ability to identify these unknown samples, which is viewed as a critical defense mechanism for secure model deployment\cite{intro:early-method,cui2022out}.

\begin{figure}[!t]
    \centering \includegraphics[width=0.98\columnwidth]{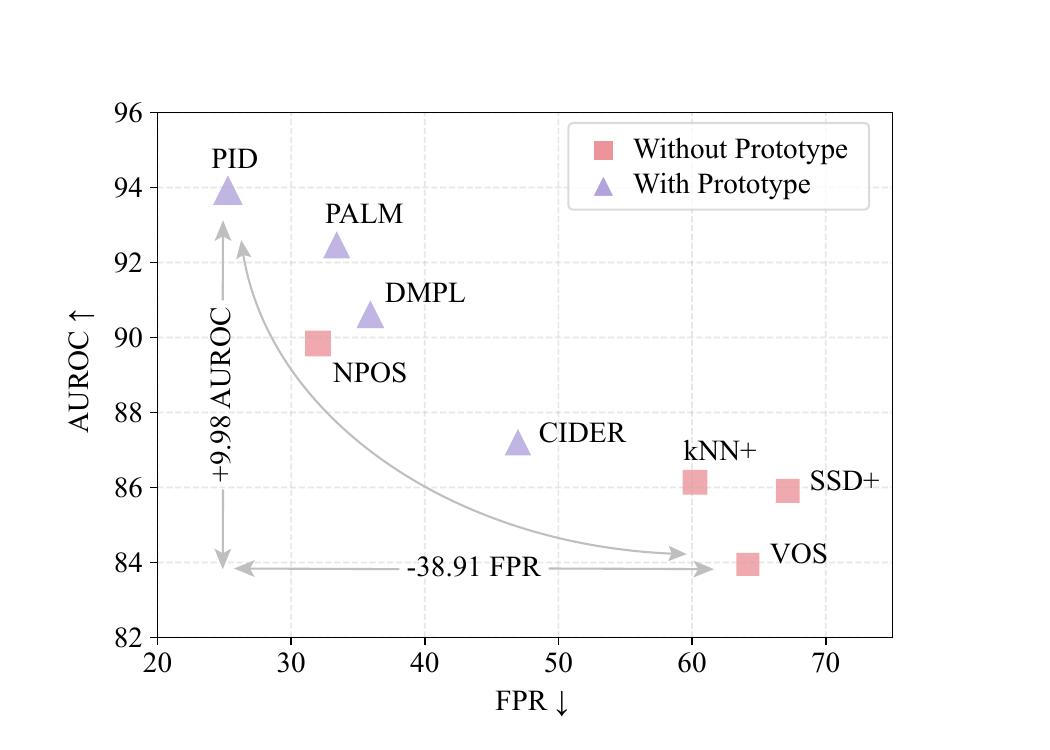} 
        \caption{
        Performance comparison of prototype-based versus non-prototype OOD detection methods. Purple triangles denote methods with prototype(e.g., PID, PALM), while pink squares denote methods without prototype(e.g., NPOS, VOS). The plot shows that our method achieves state-of-the-art (SOTA) performance, leading the prototype-based approach which, as a group, significantly improves AUROC and reduces FPR compared to non-prototype methods.
    }
    \label{fig:result}
\end{figure}

Among various OOD detection strategies, prototype-based learning methods have been established as a dominant direction\cite{du2022towards}. By these approaches, one or more representative prototypes are learned for each class of In-Distribution (ID) data within the feature space, and the assumption is made that OOD samples will lie far from these ID prototype \cite{c:30}. Early prototype learning methods tended to assign a single prototype per class\cite{peng2025distributional}, but this was recognized as an overly simplistic assumption that neglects the natural diversity inherent within a category. To address the limitations of a single prototype, features for each class have recently been modeled using multiple prototypes\cite{li2025dpu,c:62}, with the goal of achieving a finer-grained characterization of the complex substructures within ID data. However, a new constraint is introduced by these methods: a fixed number of prototypes is typically required to be manually assigned across all classes. This static assumption is found to be incapable of adapting to the inherent complexity differences across various categories. For instance, only a few prototypes might be required for a simple class, while many more are necessitated by a complex class. Currently, an effective mechanism that adaptively adjusts the number of prototypes based on data complexity is still lacking within the academic community.\newline
Inspired by the biological processes of cell birth and death\cite{gilmore2000cell}, we propose a novel method, PID (Prototype bIrth and Death), to resolve the aforementioned static assumption problem. Our method introduces two core dynamic mechanisms: prototype birth and prototype death. This dynamic prototype control mechanism is embedded within a MAP-EM framework, which is built on the prior work\cite{lu2024learning}. Through the synergistic actions of birth and death, the model's prototype count is dynamically adjusted during training, conforming to the specific data complexity of each category. By this adaptive mechanism, the model is enabled to learn more compact and better-separated ID embeddings, thereby significantly enhancing the capability for OOD sample detection. Experiments on benchmark datasets such as CIFAR-100 show that our dynamic approach significantly outperforms existing State-of-the-Art (SOTA) methods across multiple metrics, particularly on FPR95.
The main contributions of this paper can be summarized as follows:
\begin{enumerate}
    \item We propose PID (Prototype bIrth and Death), a novel dynamic prototype control mechanism for OOD detection inspired by biology, which includes prototype birth and prototype death.
    \item A variance-based overload assessment criterion for Prototype Birth and a distance ratio-based boundary score for Prototype Death are specifically designed, allowing for the self-adaptive adjustment of the prototype count.
    \item It is demonstrated that SOTA performance across multiple OOD benchmarks is achieved when this mechanism is seamlessly integrated into the MAP-EM framework.
\end{enumerate}

\section{Related Work}
\label{sec:related work}
\subsection{OOD Detection}
Motivated by the tendency of Deep Neural Networks (DNNs) to produce overconfident, erroneous predictions when processing data outside of their training distribution, Nguyen et al. first formally proposed the problem of Out-of-Distribution (OOD) detection\cite{nguyen2015deep}. To address this problem, researchers have developed various technical approaches, primarily categorized into Score-based\cite{devries2018learning}, Generative-based\cite{zhang2021understanding}, and Distance-based methods\cite{ren2021simple}.
Score-based methods focus on constructing a discrimination function using the model's output, such as confidence scores\cite{c:25, c:43}, energy values\cite{c:37, c:44, c:45}, or gradient information\cite{c:38}. Distance-based methods, on the other hand, take a different route, operating on the core assumption that OOD samples should significantly deviate from the In-Distribution (ID) data's manifold in the embedding feature space\cite{c:40, c:30,c:41,c:42}. These methods discriminate by calculating the Mahalanobis distance or the KNN distance between the test sample and the ID data and have achieved remarkable results\cite{sun2022out,yang2025oodd}. Furthermore, a few studies have begun to explore the potential of OOD detection using entirely unlabeled ID datasets\cite{zhou2017places}.
\subsection{Contrastive Learning and Prototype Learning}
Contrastive Learning captures powerful, discriminative features by forcing different views of the same input to remain close in the embedding space while being distant from other samples\cite{khosla2020supervised}. It has achieved great success in unsupervised, semi-supervised, and supervised scenarios\cite{liu2023good,lee2022contrastive,zeng2021modeling}. In particular, its fundamental properties and effectiveness in the hyperspherical space have been thoroughly investigated\cite{ming2022exploit}.
Prototypical Learning organizes the feature space by modeling the relationship between samples and pre-defined prototypes. These two strategies are often combined. For instance, Li et al. integrated prototypical learning, which additionally performs contrastive learning between samples and prototypes obtained via an offline clustering algorithm\cite{li2020prototypical}. Recent OOD detection research has further advanced Multi-Prototypical Learning to address the issue of a single prototype losing class-internal diversity\cite{c:62,xu2025hierarchical}. For example, Lu et al. proposed modeling each class as a mixture of multiple von Mises-Fisher (vMF) distributions and optimizing it with a combination of Maximum Likelihood Estimation loss and a prototypical contrastive loss\cite{banerjee2005clustering}. Jia et al. introduced a framework that combines coarse-grained and fine-grained prototypes, whose core motivation is to expose OOD samples by creating gaps between fine-grained sub-clusters. Unlike existing multi-prototypical methods that require setting a fixed number of prototypes, the proposed prototype birth and death mechanism automatically adjusts the number of prototypes based on the complexity of the class data, enhancing the model's ability to adapt to different levels of class data complexity.

\begin{figure*}[!t]
    \centering \includegraphics[width=\linewidth]{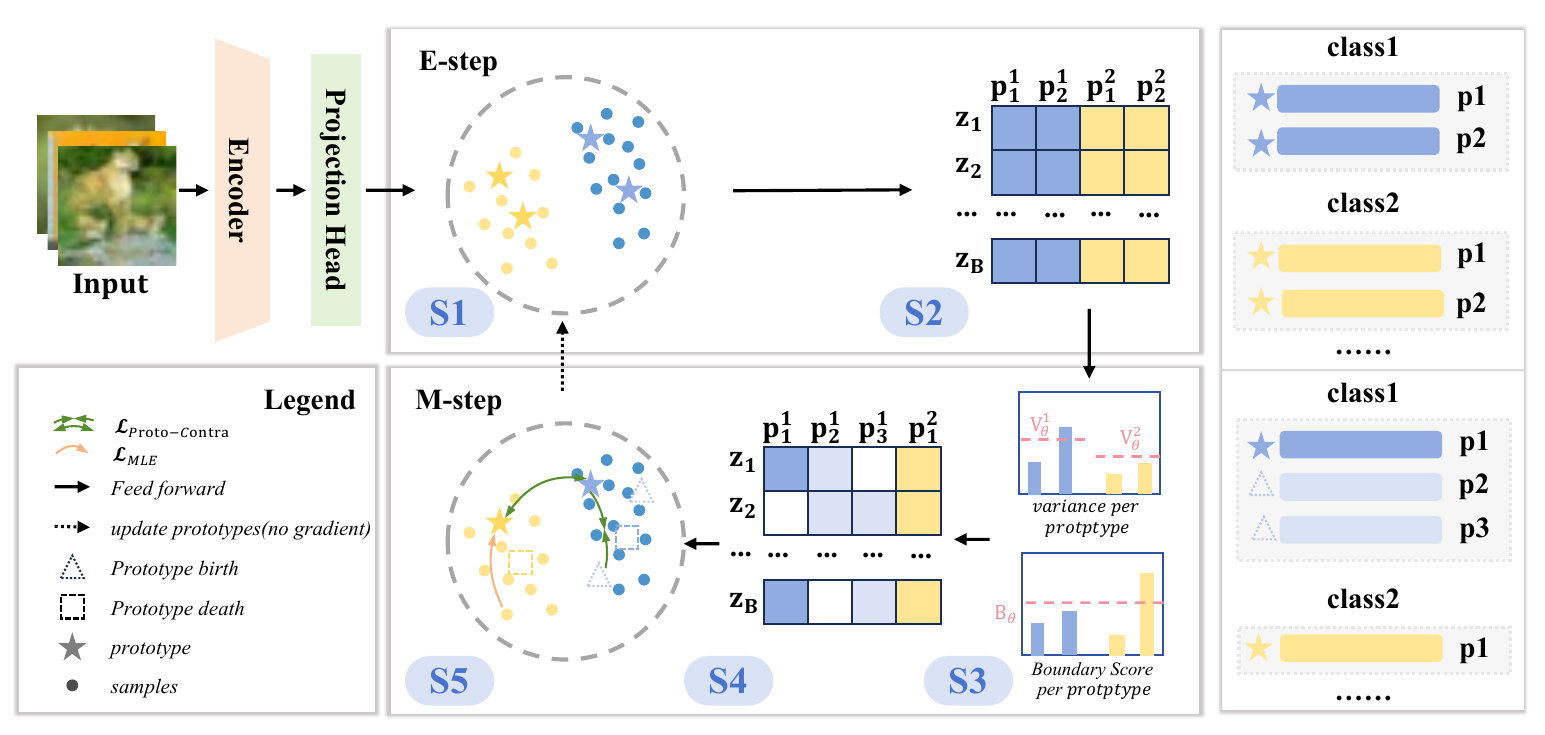} 
        \caption{
        Overview of the PID framework, illustrating one iteration of the MAP-EM algorithm. The process is divided into the E-step (top) and M-step (bottom). The E-step begins by (S1) fixing prototypes and sample representations, then (S2) computes and prunes the assignment weights. The M-step then activates: first, the Dynamic Controller (S3) evaluates the variance ($v_k^c$) and boundary score ($B_k^c$) to determine prototype birth and death. After the prototype set is updated, the assignment weights are (S4) re-calculated for the new set. Finally, (S5) performs optimization via gradient updates and non-gradient prototype updates.
    }
    \label{fig:framework_overview}
\end{figure*}

\section{Method}
\label{sec:method}

\textbf{OOD Detection Problem}
Let $\mathcal{X}$ and $\mathcal{Y}^{\mathrm{id}}=\{1,2,\ldots,C\}$ be the input space and the label space containing $C$ categories for the In-Distribution (ID) dataset, respectively. The training dataset $D^{\mathrm{id}}=\{(x_i,y_i)\}$ of the model is independently and identically distributed (i.i.d.) sampled from the joint distribution $P_{\mathcal{X}\times\mathcal{Y}^{\mathrm{id}}}$. Training and testing are assumed to be performed on the same distribution in the closed-world setting. Mathematically, the label of Out-of-Distribution (OOD) data on the test set satisfies $y\notin\mathcal{Y}^{\mathrm{id}}$, which means that $\mathcal{Y}^{\mathrm{id}}\bigcap\mathcal{Y}^{\mathrm{ood}}=\emptyset$. Let $P_{\mathrm{in}}$ denote the marginal distribution of $\mathcal{X}$ for the ID dataset in $P_{\mathcal{X}\times\mathcal{Y}^{\mathrm{id}}}$, and conversely, let $P_{\mathrm{out}}$ denote the marginal distribution of the OOD dataset. The goal of OOD detection is to determine whether a test sample $x\in\mathcal{X}$ belongs to $P_{\mathrm{in}}$ or $P_{\mathrm{out}}$ according to $S(x)<\varepsilon$, where $S(x)$ is the scoring function and $\varepsilon$ is a threshold.

\subsection{Hypersphercical Embedding Modeling}
Following prior research, we define the embedding space as a hyperspherical model. Given that the projected vector $z$ lies on the unit sphere $(\|z\|_{2}=1)$, it is natural to model it using the von Mises-Fisher (vMF) distribution, which is the natural counterpart of the Gaussian distribution on the hypersphere. Our method employs a mixture vMF distribution, motivated by the theoretical and empirical benefits of the multi-prototype assumption. The probability density function corresponding to the $k$-th prototype $p_{k}\in \mathbb{R}^d$ is defined as: $p_{\mathcal{D}}\left(z;p_k,\kappa\right)=Z_{\mathcal{D}}\left(\kappa\right)\exp{\left(\kappa p_k^{\top}z\right)}$ where $p_k$ is the mean vector of the distribution, $\kappa$ is the concentration parameter (tightness of the distribution around the mean) and $Z_{D}\left(\kappa\right)$ is the normalization constant.

Although prototype learning has shown relatively good results both theoretically and empirically, previous research relies on a fixed hyperparameter $K$, which is manually set to represent the number of prototypes assigned to each class.This approach accounts for intra-class diversity, but the hyperparameter $K$ does not assign an appropriate number of prototypes for every class, leading to redundancy in prototype representation capability in simple classes and insufficiency in complex classes.Therefore, our method assigns $K_c$ prototypes $P^c=\{p_k^c\}_{k=1}^{K_c}$ to each class $c$ to characterize the mixture vMF distribution, where the value of $K_c$ can change dynamically during the training process. Each sample representation $z_i$ is also assigned to the prototypes via assignment weights $w_i^c\in \mathbb{R}^{K_c}$. Therefore, the probability density function of sample $z_i$ belonging to class $c$ is defined as:
\begin{equation}
  p\left(z_i\middle|\ y_i=c;P^c,w_i^c,\kappa\right)=\sum_{k=1}^{K_c}{w_{i,k}^cZ_{\mathcal{D}}\left(\kappa\right)}\exp{\left(\kappa p_k^{c\top}z_i\right)}
  \label{eq:probability density function of sample}
\end{equation}
where $w_{i,k}^c$ denotes the $k$-th element in $w_i^c$. We use the same assignment weight calculation method as in prior research.Based on the above equation and Bayes' theorem, the posterior probability of an embedding $z_i$ belonging to class $c$ can be derived as:
\begin{equation}
  p\left(y_i=c\middle| z_i;\{P^j,w_i^j\}_j,\tau\right)=\frac{\sum_{k=1}^{K_c}{w_{i,k}^c\exp{\left(p_k^{c\top}z_i/\tau\right)}}}{\sum_{j=1}^{C}\sum_{k^{\prime}=1}^{K_j}{w_{i,k^{\prime}}^j\exp{\left(p_{k^{\prime}}^{j\top}z_i/\tau\right)}}}
  \label{eq:the posterior probability}
\end{equation}
where $\tau=1/\kappa$ is the temperature coefficient.

\subsection{Dynamic Prototype Management based on the MAP-EM Framework}
The prototype assignment is treated as a latent variable $W$. The network parameters $\Theta_{net} = \{\theta,\phi\}$ and prototype representations $P^c=\{p_k\}_{k=1}^{K_c}$ are considered the model parameters $\Theta = \{\Theta_{net}, P\}$. The objective is to maximize the log-likelihood of the observed data $Z$, denoted as $p(Z|\Theta)$. However, direct optimization is challenging due to the presence of the latent variable $W$. Therefore, a variant of the Expectation-Maximization (EM) algorithm, the Maximum A Posteriori Expectation-Maximization (MAP-EM) algorithm, is adopted for alternating optimization, inspired by Prototype Contrastive Learning. Unlike standard EM, which aims to maximize the expected log-likelihood, the M-step of MAP-EM maximizes the expected log posterior probability, specifically $Q(\Theta | \Theta^{(t)}) + \log p(\Theta)$. This formulation facilitates the introduction of a prior $p(\Theta)$ to regularize the model parameters, thereby guiding the model toward a more optimal structure for OOD detection.

\subsubsection{E-Step: Prototype Assignment Estimation and Sparsification}
In the E-step (Expectation), the current model parameters $\Theta^{(t)} = \{\theta^{(t)}, \phi^{(t)}, P^{(t)}\}$ and the number of prototypes for each class $\{K_c^{(t)}\}_{c=1}^C$ are fixed. The goal of this step is to compute the posterior probability distribution of the latent variable $W$, $p(W | Z, \Theta^{(t)})$, which specifically involves estimating the assignment weight $w_{i,k}^c$ for each sample $z_i$ to the prototype $p_k^c$ of its corresponding class $c$.This process is divided into two sub-steps: 

\noindent\textbf{Globally Optimal Assignment}
First, the hyperspherical embeddings $Z$ for all samples in the current batch are computed using the fixed encoder $f_{\theta^{(t)}}$ and projection head $g_{\phi^{(t)}}$. For each class $c$, the similarity matrix between its samples $Z^c$ and its prototypes $P^c$ is calculated. The Sinkhorn-Knopp algorithm, as utilized in PALM, is employed to compute the optimal soft assignment matrix $W^c$. This iterative normalization process globally balances the assignments, ensuring uniform utilization of the prototypes. 

\noindent\textbf{Assignment Pruning and Sparsification}
The matrix $W^c$ obtained via Sinkhorn is typically dense. To reinforce the association between a sample and its truly nearest prototypes, a sparsification pruning step is introduced. A Top-k filtering is applied to the assignment vector $w_i^c$ (the column of $W^c$) for each sample $z_i$. Only the $K$ most relevant prototypes retain non-zero weights (which are then re-normalized), while the weights for all other prototypes are set to zero.
This strategy ensures both assignment diversity and robustness via sparsification, providing a clearer and more stable signal for prototype updates in the subsequent M-step.

\subsubsection{M-Step: Model Parameter and Prototype Optimization}
In the M-step (Maximization), the sparse assignment weights $W^{(t+1)}$ obtained from the E-step are fixed, and the model parameters $\Theta$ are updated to maximize the expected log posterior probability:
\begin{equation}
  \Theta^{(t+1)} = \arg \max_{\Theta} \left[ Q(\Theta | \Theta^{(t)}) + \log p(\Theta) \right]
  \label{eq:expected log posterior probability}
\end{equation}
As derived in Appendix A, this maximization objective is equivalent to minimizing a joint loss function $\mathcal{L}$. This step comprises three core components: 

\noindent\textbf{Encoder Parameter Optimization}
The parameters of the encoder $f_{\theta}$ and the projection head $g_{\phi}$, denoted as $\Theta_{net}$, are optimized via gradient descent. The objective is to minimize the following joint loss function:
\begin{equation}
  \mathcal{L} = \mathcal{L}_{MLE} + \lambda \mathcal{L}_{proto-contra}
  \label{eq:joint loss function}
\end{equation}
$\mathcal{L}_{MLE}$ corresponds to the main term of the negative expected log-likelihood $(-Q(\Theta | \Theta^{(t)}))$. It utilizes the assignment weights $W$ to pull each sample $z_i$ closer to its assigned prototype clusters by minimizing the Negative Log-Likelihood (NLL):
\begin{equation}
  \mathcal{L}_{MLE} = -\frac{1}{N} \sum_{i=1}^{N} \log \frac{\sum_{k=1}^{K_{y_i}}{w_{i,k}^{y_i} \exp{\left(p_k^{y_i \top} z_i / \tau \right)}}}{\sum_{j=1}^{C}\sum_{k'=1}^{K_j}{w_{i,k'}^j \exp{\left(p_{k'}^{j \top} z_i / \tau \right)}}}
  \label{eq:mle loss}
\end{equation}
$\mathcal{L}_{proto-contra}$ corresponds to the negative log-prior of the prototype representations $(-\log p(\Theta))$. The prototype contrastive loss is treated as a structural prior imposed on the prototypes $P$, favoring a configuration that promotes intra-class compactness and inter-class separation. This InfoNCE-style loss can be written in a simplified form:
\begin{equation}
    \mathcal{L}{proto-contra} = -\frac{1}{\sum{c=1}^C K_c} \sum_{c=1}^{C} \sum_{k=1}^{K_c} \log \left( s_k^c \right)
    \label{eq:prototye contrastive loss}
\end{equation}
Here, $s_k^c$ is the score for prototype $p_k^c$, defined as the ratio between positive (intra-class) similarities and all similarities (positive and negative). For brevity, we define the indicator $\mathbb{I}_{k, k''}^c = \mathbb{I}(j \neq c \text{ or } k'' \neq k)$. The score $s_k^c$ is then:
\begin{equation}
    s_k^c = \frac{\sum_{k'=1, k' \neq k}^{K_c} \exp{\left( p_k^{c \top} p_{k'}^c / \tau_p \right)}}{\sum_{j=1}^{C} \sum_{k''=1}^{K_j} \mathbb{I}{k, k''}^c \exp{\left( p_k^{c \top} p{k''}^j / \tau_p \right)}}
    \label{eq:prototye contrastive loss_score}
\end{equation}

By minimizing $\mathcal{L}$, $\theta$ and $\phi$ are updated, driving the embedding space to evolve in a direction that simultaneously fits the data ($\mathcal{L}_{MLE}$) and satisfies the structural requirements necessary for OOD detection ($\mathcal{L}_{proto-contra}$). 

\noindent\textbf{Dynamic Prototype Controller}
This mechanism represents the core innovation of the method. After parameter optimization, the number of prototypes $K_c$ for each category is dynamically adjusted.
\begin{itemize}
    \item \textbf{Prototype Birth} The overload status of each prototype $p_k^c$ is continuously monitored, quantified by the intrinsic variance $v_k^c$ of the sample cluster it represents. This variance is defined as the average squared Euclidean distance from the set of samples belonging to the prototype ($Z_k^c$) to their mean ($\mu_k^c$):
    \begin{equation}
        v_k^c = \frac{1}{|Z_k^c|} \sum_{z_i \in Z_k^c} ||z_i - \mu_k^c||_2^2
        \label{eq:variance}
    \end{equation}
    Concurrently, the average variance of all $K_c$ prototypes within class $c$ is computed:
    \begin{equation}
        v_{avg}^c = \frac{1}{K_c} \sum_{j=1}^{K_c} v_j^c
        \label{eq:average variance}
    \end{equation}
    When a prototype's variance $v_k^c$ consistently exceeds a dynamic threshold established by $\lambda^c$ times the intra-class average variance factor($\lambda^c > 1$), i.e., $v_k^c > \lambda^c \cdot v_{avg}^c$, it indicates that the sub-cluster represented by this prototype is too dispersed and its representational capacity is insufficient. For simplicity, the variance threshold factor $\lambda^c$ is set to be the same across all classes $c$ during implementation. At this point, the birth mechanism is triggered: the overloaded prototype is split into two new sub-prototypes along the principal component (PCA) direction of its sample distribution.
\end{itemize}
\begin{itemize}
    \item \textbf{Prototype Death} The redundancy status of prototypes is evaluated simultaneously. A boundary score $B_k^c$ is designed to quantify the ratio of the distance between the prototype and its nearest inter-class prototype to the distance between it and its nearest intra-class prototype:
    \begin{equation}
        B_k^c = \frac{\min_{j \neq c, k'} \left( 1 - p_k^{c \top} p_{k'}^j \right)}{\min_{k' \neq k} \left( 1 - p_k^{c \top} p_{k'}^c \right)}
        \label{eq:boundary score}
    \end{equation}
    A low score $B_k^c$ suggests that the prototype is either too close to an inter-class prototype or too far from an intra-class prototype, indicating its location in a blurred region of the class decision boundary, and thus contributing minimally to class discriminability. For simplicity, the threshold used for the boundary score $B_k^c$ is set to be the same across all prototypes $p_k^c$. These prototypes are flagged for the death mechanism and subsequently removed.
\end{itemize}
\begin{itemize}
    \item \textbf{Stability Mechanism} We employ a phasing strategy for dynamic prototype adjustment, activating it only after features stabilize to ensure effectiveness and deactivating it near the end of training to ensure final convergence.Additionally, to prevent the number of prototypes $K_c$ from drastically oscillating during this active phase, which could destabilize training, a global cooldown period is introduced. After any prototype birth or death event, the dynamic controller pauses its operation for a certain number of epochs, allowing sufficient time for the network parameters and sample assignments to adapt to the new prototype structure.
\end{itemize}

\noindent\textbf{Prototype Representation Update} The prototypes are non-parametric, and their positions are thus not learned via gradient descent. Instead, Exponential Moving Average (EMA) is utilized for smooth updates, which corresponds to the non-gradient optimization of $P$ in the M-step:
\begin{equation}
    p_k^c \leftarrow \text{\textit{Normalize}} \left( \alpha p_k^c + (1-\alpha) \sum_{i=1}^{B} \mathbb{I}(y_i=c) w_{i,k}^c z_i \right)
    \label{eq:ema update}
\end{equation}
where $\text{\textit{Normalize}}(v) = v / ||v||_2$, $\mathbb{I}(\cdot)$ is the indicator function, and $\alpha$ is the EMA momentum coefficient. The use of EMA offers two main benefits: (1) It acts as a momentum term, resulting in smoother prototype movement and filtering noise introduced by mini-batch sampling; (2) It ensures the asynchronous updating of prototypes and the encoder, which is crucial for preventing the EM algorithm from converging to trivial solutions. 
Through the alternating iteration of the E-step and M-step, the model not only learns a compact and well-separated embedding space but also automatically discovers the optimal number of prototypes $K_c$ required for each category, adapting to the intrinsic complexity of different classes.

\section{Experiments}
\label{sec:experiments}

\begin{table*}[!ht]
    \centering
    \caption{OOD detection performance on methods trained on \textbf{labeled CIFAR-100} as ID dataset using backbone network of ResNet-34. ↓ means smaller values are better and ↑ means larger values are better. \textbf{Bold} numbers indicate superior results and \underline{underline} numbers indicate the second-best results.}
    \label{tab:ood_cifar100}
    \small % Make the table font slightly smaller if needed
    \setlength{\tabcolsep}{5pt} % Adjust column spacing if needed

    \resizebox{\textwidth}{!}{%
    
    \begin{tabular}{@{}l c c c c c c c c c c c c@{}}
        \toprule
        \multirow{3}{*}{Methods} & \multicolumn{10}{c}{OOD Datasets} & \multicolumn{2}{c}{Average} \\
        \cmidrule(lr){2-11} \cmidrule(lr){12-13}
        & \multicolumn{2}{c}{SVHN} & \multicolumn{2}{c}{Places365} & \multicolumn{2}{c}{LSUN} & \multicolumn{2}{c}{iSUN} & \multicolumn{2}{c}{Textures} & \multicolumn{2}{c}{} \\
        & FPR↓ & AUROC↑ & FPR↓ & AUROC↑ & FPR↓ & AUROC↑ & FPR↓ & AUROC↑ & FPR↓ & AUROC↑ & FPR↓ & AUROC↑ \\
        \midrule
        MSP      & 78.89          & 79.80          & 84.38          & 74.21       & 83.47          & 75.28          & 84.61          & 74.51          & 86.51          & 72.53          & 83.57          & 74.62          \\
        Vim      & 73.42          & 84.62          & 85.34          & 69.34          & 86.96          & 69.74          & 85.35          & 73.16          & 74.56          & 76.23          & 81.13          & 74.62          \\
        ODIN     & 70.16          & 84.88          & 82.16          & 75.19          & 76.36          & 80.10          & 79.54          & 79.16          & 85.28          & 75.23          & 78.70          & 78.91          \\
        Energy   & 66.91          & 85.25          & 81.41          & 76.37          & 59.77          & 86.69          & 66.52          & 84.49          & 79.01          & 79.96     & 70.72          & 82.55          \\
        VOS      & 43.24          & 82.80          & 76.85          & 78.63          & 73.61          & 84.69          & 69.65          & 86.32          & 57.57          & 87.31          & 64.18          & 83.95          \\
        CSI      & 44.53          & 92.65          & 79.08          & 76.27          & 75.58          & 83.78          & 76.62       & 84.98          & 61.61          & 86.47          & 67.48     & 84.83          \\
        SSD+     & 31.19          & 94.19          & 77.74          & 79.90          & 79.39          & 85.18          & 80.85          & 84.08          & 66.63          & 86.18          & 67.16          & 85.91          \\
        kNN+     & 39.23          & 92.78          & 80.74          & 77.58          & 48.99          & 89.30          & 74.99          & 82.69          & 57.15          & 88.35          & 60.22          & 86.14          \\
        NPOS     & 10.62          & 97.49          & 67.96          & 78.81          & 20.61          & 92.61          & \underline{35.94} & 88.94          & \textbf{24.92} & 91.35          & \underline{32.01} & 89.84          \\
        CIDER    & 22.95          & 95.17          & 79.81          & 73.59          & 16.19         & 96.32         & 71.96          & 80.54     & 43.94         & 90.42         & 46.97          & 87.21        \\
        PALM     & \underline{3.03} & \underline{99.23} & \underline{67.80} & \textbf{82.62} & \underline{10.58} & \underline{97.70} & 41.56          & \underline{91.36} & 44.06          & \underline{91.43} & 33.41          & \underline{92.47} \\
        DMPL    & 16.97  & 96.24 & 73.66          & 79.55          & 13.10          & 96.85        & 40.83          & 89.51          & 35.09      & 90.93         & 35.93          & 90.62          \\
        \midrule % Add a midrule before your new methods for visual separation
        % --- Your methods below this line ---
        \textbf{PID} & \textbf{2.08}$^{\pm 1.07}$ & \textbf{99.56}$^{\pm 0.23}$ & \textbf{65.75}$^{\pm 1.91}$ & \underline{82.30}$^{\pm 2.02}$ & \textbf{8.78}$^{\pm 2.46}$ & \textbf{98.02}$^{\pm 0.83}$ & \textbf{18.48}$^{\pm 5.27}$ & \textbf{96.19}$^{\pm 1.38}$ & \underline{31.26}$^{\pm 3.52}$ & \textbf{93.59}$^{\pm 0.44}$ & \textbf{25.27}$^{\pm 1.43}$ & \textbf{93.93}$^{\pm 0.53}$ \\
       % --- Add more methods as needed ---
        \bottomrule
    \end{tabular}
    
    }%
\end{table*}

\begin{table*}[!ht]
    \centering
    \caption{\textit{Near-OOD detection} performance on methods trained on \textbf{labeled CIFAR-100} as ID dataset using backbone network of ResNet-34. ↓ means smaller values are better and ↑ means larger values are better. \textbf{Bold} numbers indicate superior results and \underline{underline} numbers indicate the second-best results.}
    \label{tab:near_ood_cifar100}
    \scriptsize
    
    \setlength{\tabcolsep}{0pt}
    \begin{tabular*}{\textwidth}{@{}l @{\extracolsep{\fill}} c c c c c c c c c c@{}}
        \toprule
        \multirow{3}{*}{Methods} & \multicolumn{8}{c}{OOD Datasets} & \multicolumn{2}{c}{Average} \\
        \cmidrule(lr){2-9} \cmidrule(lr){10-11}
        & \multicolumn{2}{c}{LSUN-F} & \multicolumn{2}{c}{ImageNet-F} & \multicolumn{2}{c}{ImageNet-R} & \multicolumn{2}{c}{CIFAR-10} & \multicolumn{2}{c}{} \\
        & FPR↓ & AUROC↑ & FPR↓ & AUROC↑ & FPR↓ & AUROC↑ & FPR↓ & AUROC↑ & FPR↓ & AUROC↑ \\
        \midrule
        MSP     & 88.24         & 69.21         & 86.33         & 70.74         & 86.32         & 72.88         & 88.06         & \textbf{76.30} & 87.24         & 72.28         \\
        Energy  & 87.17         & 72.20         & 78.99         & 76.40         & 80.93         & 80.60         & 86.47         & 70.50         & 84.21         & 74.93         \\
        SSD+    & 83.36         & \textbf{76.63} & 76.73         & 79.78         & 83.67         & 81.09         & 85.16         & 73.70         & 82.23         & 77.80         \\
        kNN+    & 84.96         & 75.37         & 75.52         & 79.95         & 68.49         & 84.91         & \textbf{84.12} & \underline{75.91} & 78.27         & 79.04         \\
        CIDER   & 90.94         & 70.31         & 78.83         & 77.53         & 56.89         & 87.62         & \underline{84.87} & 73.30         & 77.88         & 77.19         \\
        PALM    & \underline{79.18} & 75.51         & \textbf{68.48} & \textbf{80.54} & \underline{28.68} & \underline{92.91} & 89.59         & 70.21         & \underline{66.48} & \underline{79.79} \\
        \midrule
        \textbf{PID} & \textbf{76.17} & \underline{75.88} & \underline{68.89} & \underline{80.44} & \textbf{19.28} & \textbf{96.00} & 90.02         & 67.71         & \textbf{63.59} & \textbf{80.01} \\
        \bottomrule
    \end{tabular*}
\end{table*}

\subsection{Experimental Setup}

\textbf{Datasets and Training Details} Our core experiments are conducted on CIFAR-100 \cite{krizhevsky2009learning}, using ResNet-34 as the backbone network. To validate the generalization ability of our method, we also conduct experiments on CIFAR-10 and the larger-scale ImageNet-100 dataset, using their respective backbones (ResNet-18 and ResNet-50). Our OOD test sets cover far-OOD scenarios (SVHN \cite{netzer2011reading}, Textures \cite{cimpoi2014describing}, LSUN \cite{yu2015lsun}, iSUN \cite{xu2015turkergaze}, and Places365 \cite{zhou2017places}) as well as more challenging near-OOD scenarios (LSUN-FIX, ImageNet-FIX \cite{beyer2020we}, ImageNet-RESIZE \cite{c:43}, and CIFAR-10 \cite{krizhevsky2009learning}). Unless otherwise specified, all experiments in this section use the CIFAR-100 dataset and the ResNet-34 backbone. Following standard practice, we attach an MLP projection head after the backbone, which embeds features onto a 128-dimensional unit hypersphere. All models are trained for 500 epochs using an SGD optimizer with momentum. To ensure fairness, we share the same hyperparameters with baseline methods whenever possible. Additionally, for the proposed dynamic prototype control mechanism, we set the intra-class average variance factor for prototype birth to 2.0, the boundary score threshold for prototype death to 2.5, the global cooldown period to 5 epochs, and the birth and death periods to 50 epochs each.
\newline\textbf{OOD Detection Scoring Function} Given that our method aims to learn the true data distribution, we select a widely-used distance-based OOD detection method, the Mahalanobis score \cite{mclachlan1999mahalanobis}. Following standard practice \cite{c:42,sun2022out}, we utilize the feature embeddings from the penultimate layer for distance metric calculation.
\newline\textbf{Evaluation Metrics} To demonstrate the effectiveness of PID, we report three commonly used evaluation metrics: (1) The False Positive Rate (FPR) of OOD samples when the True Positive Rate (TPR) of ID samples is 95\%; (2) the Area Under the Receiver Operating Characteristic curve (AUROC); and (3) the Area Under the Precision-Recall curve (AUPR).

\subsection{Main Results}
\begin{figure*}[!t]
  \centering
  \includegraphics[width=0.98\textwidth]{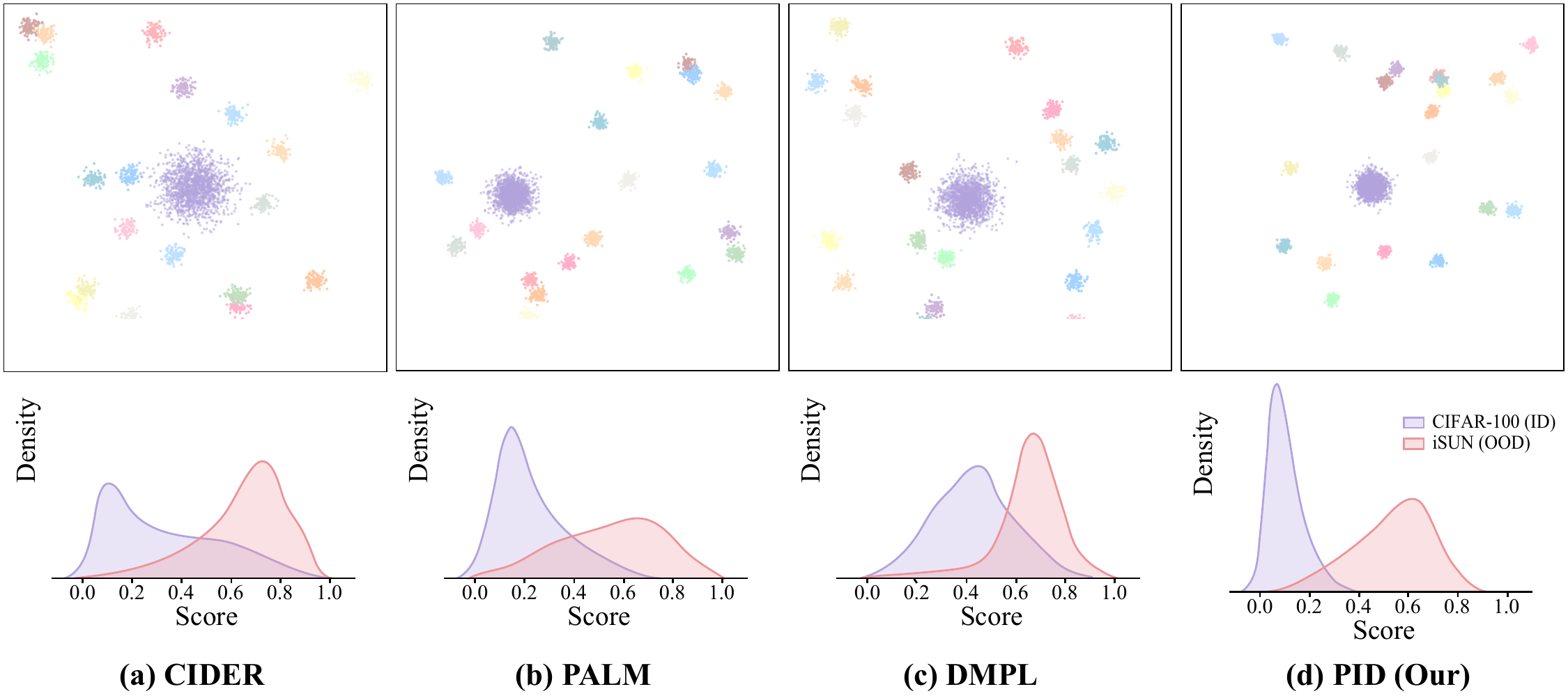}
  \caption{UMAP visualization of the first 20 subclasses of ID (CIFAR-100) and all OOD (iSUN) samples plotted to the same embedding space for methods including (a) CIDER (b) PALM (c) DMPL and (d) PID (Our). The scores are obtained by scaling the distance metrics used by each method to [0, 1] for visualization.}
  \label{fig:umap_pid_comparison}
\end{figure*}

PID outperforms state-of-the-art (SOTA) methods on all OOD benchmarks by a significant margin. As shown in Table \ref{tab:ood_cifar100}, under the standard setting on CIFAR-100, our method is compared against a suite of highly competitive baselines, including MSP \cite{c:25}, Vim \cite{wang2022vim}, ODIN \cite{c:43}, Energy \cite{c:37}, VOS \cite{du2022towards}, CSI \cite{c:40}, SSD+ \cite{c:42}, kNN+ \cite{sun2022out}, NPOS \cite{c:30}, CIDER \cite{ming2022exploit}, PALM \cite{lu2024learning}, and DMPL\cite{c:62}. PID achieves a comprehensive lead across all five OOD datasets and both primary evaluation metrics. Most notably, compared to the closest competitor in terms of overall performance, PID achieves a remarkable 24.36\% relative reduction in average FPR while simultaneously boosting the average AUROC from 92.47\% to 93.93\%. This superior performance is primarily attributed to our proposed dynamic prototype control mechanism. Unlike the static prototypes or dependency modeling employed by methods like PALM or CIDER, our dynamic mechanism enables the model to learn a more refined embedding space structure that better fits the true data manifold. Furthermore, even on notoriously challenging OOD datasets such as Places365, PID still demonstrates best-in-class performance, significantly outperforming all other baseline methods.We further provide a qualitative visualization in Figure \ref{fig:umap_pid_comparison}. The UMAP plot (top) visually confirms that our PID (d) learns significantly more compact intra-class clusters and achieves better separation of OOD samples (purple) compared to baselines (a-c). Consequently, the score density plot (bottom) shows that PID achieves the cleanest separation between the ID (blue) and OOD (red) distributions.

\begin{table}[htbp]
    \centering
    \caption{Ablation study on the effectiveness of Birth and Death Mechanisms of our method on CIFAR-100. All metrics are averaged (\%).}
    \label{tab:birth_and_dirth_ablation}
    \small
    \begin{tabular*}{\columnwidth}{@{}l @{\extracolsep{\fill}} c c c@{}}
    \toprule
    Components & FPR@95$\downarrow$ & AUROC$\uparrow$ & AUPR$\uparrow$ \\ 
    \midrule
    w/o birth\&death & 43.04 & 89.82 & 91.09 \\
    w/o birth & 41.20 & 90.43 & 91.28 \\
    w/o death & 34.42 & 91.28 & 91.66 \\
    Ours & \textbf{25.27} & \textbf{93.93} & \textbf{94.35} \\ 
    \bottomrule
    \end{tabular*}
\end{table}

\subsection{Near-OOD Detection Results}
PID successfully extends its superior performance from far-OOD scenarios to the more challenging Near-OOD benchmarks. As shown in Table \ref{tab:near_ood_cifar100}, our method outperforms all baselines, achieving SOTA performance with an average FPR of 63.59\% and an average AUROC of 80.01\%. PID's superiority is primarily demonstrated by its performance on the ImageNet-R dataset. This strongly demonstrates that our dynamic prototype control mechanism possesses exceptional robustness against extreme style and texture variations, a notable limitation of static prototype methods. Furthermore, despite its suboptimal performance on unique near-OOD challenges like CIFAR-10, PID nonetheless secures the overall SOTA position for near-OOD detection.

\subsection{Ablation Study}

\begin{table}[t]
    \centering
    \caption{Ablation study on the intra-class average variance factor $\lambda^c$ of Birth Mechanism on CIFAR-100. All metrics are averaged (\%).}
    \label{tab:avg_variance_factor}
    \small
    \begin{tabular*}{\columnwidth}{@{}l @{\extracolsep{\fill}} c c c@{}}
    \toprule
    $\lambda^c$ & {FPR@95$\downarrow$} & {AUROC$\uparrow$} & {AUPR$\uparrow$} \\ 
    \midrule
    1.7 & 34.71 & 91.94 & 92.80 \\
    1.8 & 33.44 & 91.76 & 92.33 \\
    1.9 & 36.70 & 91.72 & 92.62 \\
    2.0 & \textbf{25.27} & \textbf{93.93} & \textbf{94.35} \\ 
    2.1 & 38.56 & 90.52 & 90.92 \\
    2.2 & 42.17 & 89.48 & 90.69 \\
    \bottomrule
    \end{tabular*}
\end{table}

\subsubsection{Effectiveness of Birth and Death Mechanisms}
To validate the necessity of the combined use of the birth and death mechanisms in our proposed method, we conducted a series of rigorous ablation studies on the CIFAR-100 dataset, as shown in Table \ref{tab:birth_and_dirth_ablation}. While keeping all other parameter configurations identical, we evaluated the model performance when: (1) removing both birth and death mechanisms (w/o birth\&death); (2) removing only the birth mechanism (w/o birth); and (3) removing only the death mechanism (w/o death).

The experimental results clearly indicate that removing either mechanism individually (w/o birth or w/o death) leads to a performance drop, while removing both mechanisms simultaneously (w/o birth\&death) causes a significant performance degradation, with the FPR@95 increasing to $43.04\%$. In contrast, the complete model utilizing both birth and death mechanisms (ours) achieved the best performance across all evaluation metrics. This result fully demonstrates the necessity of both the birth and death components for enhancing the model's OOD detection performance and highlights the critical synergistic effect between them.

\begin{table}[t]
    \centering
    \caption{Ablation study on the boundary score $B_k^c$ of Death Mechanism on CIFAR-100. All metrics are averaged (\%).}
    \label{tab:boundary_score}
    \small
    \begin{tabular*}{\columnwidth}{@{}l @{\extracolsep{\fill}} c c c@{}}
    \toprule
    $B_k^c$ & {FPR@95$\downarrow$} & {AUROC$\uparrow$} & {AUPR$\uparrow$} \\ 
    \midrule
    2.3 & 29.23 & 93.59 & 94.19 \\
    2.4 & 31.12 & 93.23 & 94.11 \\
    2.5 & \textbf{25.27} & \textbf{93.93} & \textbf{94.35} \\ 
    2.6 & 32.85 & 92.30 & 93.20  \\
    2.7 & 28.98 & 93.52 & 94.23 \\
    \bottomrule
    \end{tabular*}
\end{table}

\begin{figure}[b]
  \centering 
  \begin{subfigure}[b]{0.48\columnwidth}
    \includegraphics[width=\linewidth]{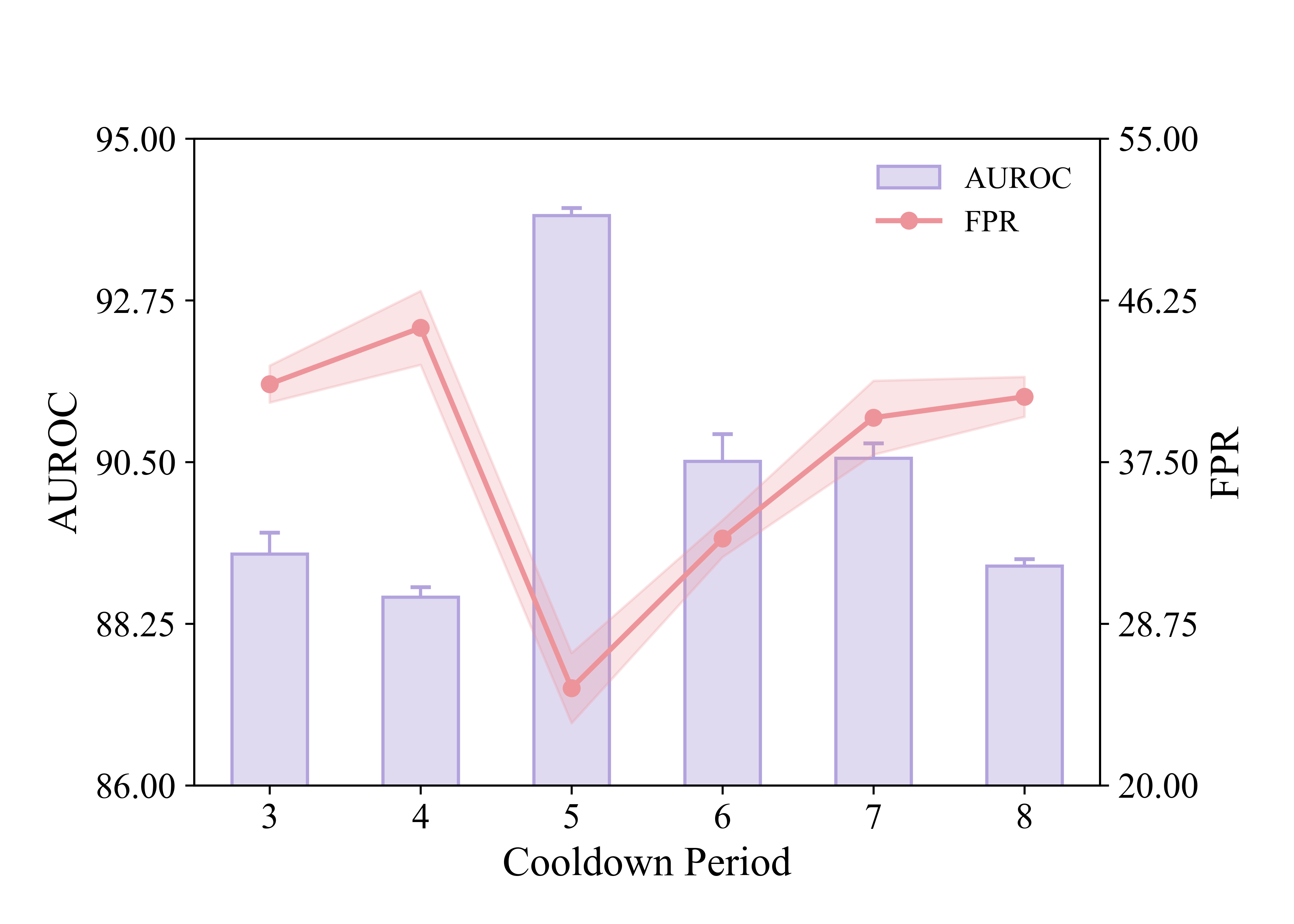}
    \caption{ }
    \label{fig:cooldown}
  \end{subfigure}%
  \hfill %这一行和上面一行下面一行之间绝对不能有空行
  \begin{subfigure}[b]{0.48\columnwidth} 
    \includegraphics[width=\linewidth]{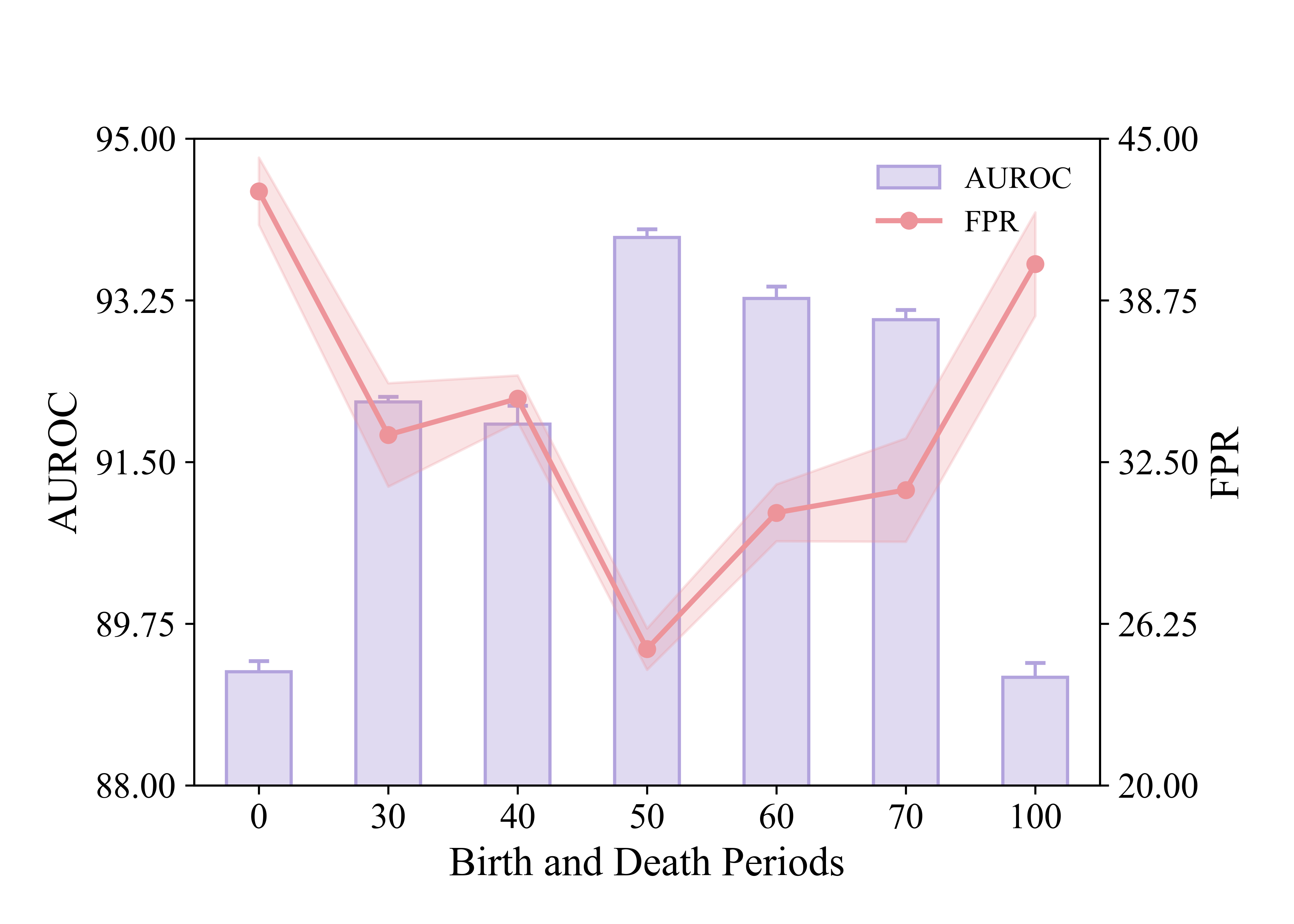}
    \caption{ }
    \label{fig:bd_period}
  \end{subfigure}
  \caption{Ablation study on the training hyperparameters of our dynamic mechanisms on CIFAR-100. (a) Sensitivity analysis of the cooldown period. (b) Sensitivity analysis of the birth and death period.}
  \label{fig:train_period_hyperparameters}
\end{figure}

\subsubsection{Key Hyperparameter Sensitivity Analysis}
We analyze the variance factor $\lambda^c$ of the Birth Mechanism and the boundary score $B_k^c$ of the Death Mechanism. As shown in Table \ref{tab:avg_variance_factor} and \ref{tab:boundary_score}, the model performance exhibits high sensitivity to $\lambda^c$. Optimal performance is achieved only at $\lambda^c = 2.0$, with a slight deviation of $\pm 0.1$ causing a sharp performance decline, confirming that $\lambda^c$ requires precise tuning.In contrast, $B_k^c$ demonstrates significant robustness. Although $B_k^c = 2.5$ is the optimal value, the model's performance curve is relatively flat across the tested range, indicating that our method is not heavily dependent on the precise selection of $B_k^c$.

\begin{figure}[!t]
    \centering \includegraphics[width=0.92\columnwidth]{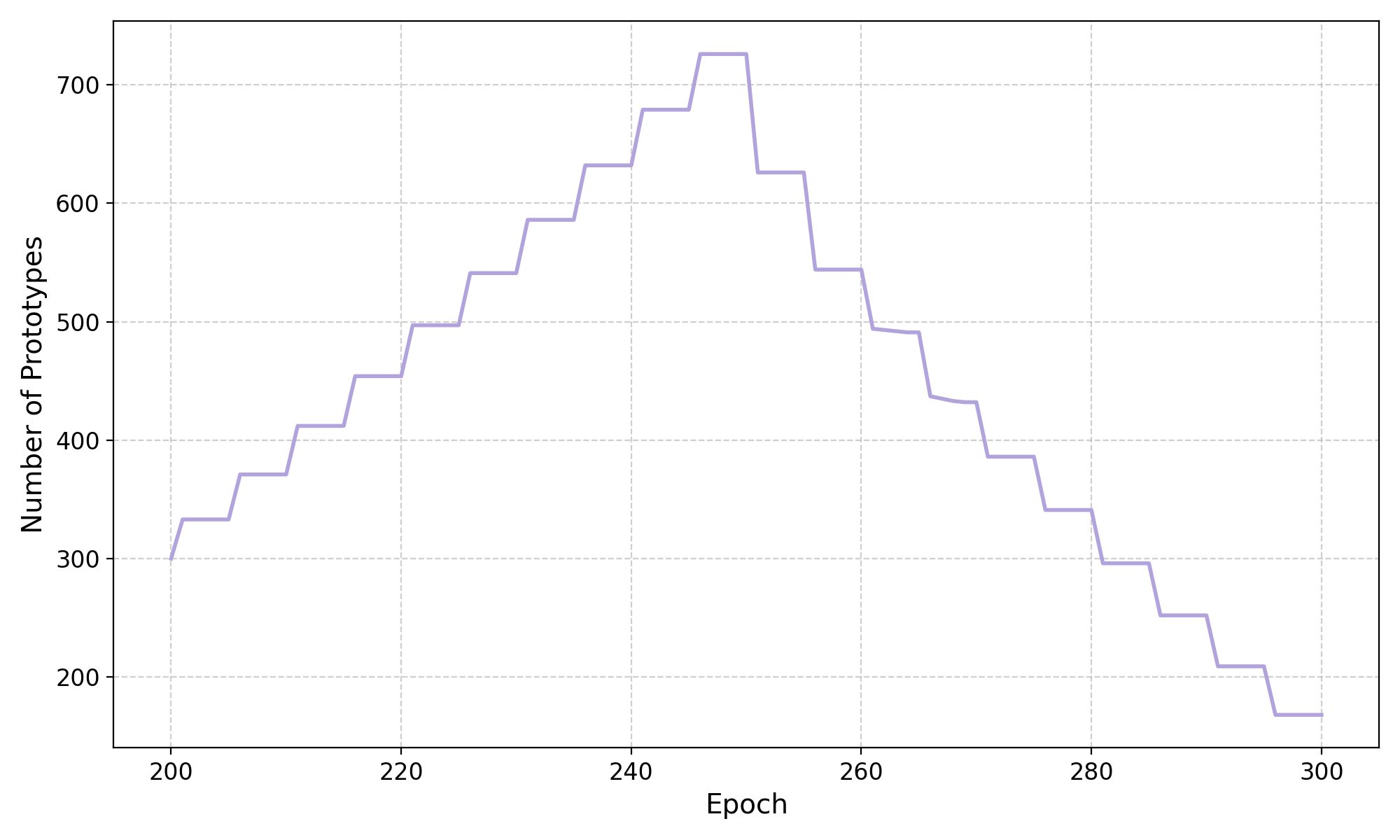} 
        \caption{
        The dynamic evolution of the total number of prototypes during training (Epoch 200-300).
    }
    \label{fig:number_change}
\end{figure}

\subsubsection{Analysis of Training Period Hyperparameters}
We also investigate the periodic hyperparameters of the dynamic mechanism (Figure \ref{fig:train_period_hyperparameters}). The results highlight that the Cooldown Period (Figure \ref{fig:cooldown}) is an extremely sensitive parameter. Deviating from the optimum causes a cliff-like drop in performance, suggesting that an improper cooldown setting severely destabilizes prototypes during initial training.
The Birth and Death Periods (Figure \ref{fig:bd_period}) are far more robust. Its performance curve is smoother, maintaining a high AUROC and low FPR within a wide range (e.g., 30 to 70 epochs) around the 50-epoch optimum.

\subsubsection{Analysis of Dynamic Prototype Number}
As shown in Figure \ref{fig:number_change}, the evolution of the prototype count during epoch 200 to 300 clearly illustrates the two stages of our dynamic mechanism. From epoch 200 to 250, the Birth Mechanism is activated, and the number of prototypes steadily increases, reaching a peak of 726. This stage corresponds to the full exploration of the ID data's feature space to ensure the completeness of the representation. Subsequently, from epoch 250 to 300, the Death Mechanism is activated, causing the total number of prototypes to systematically decrease, ultimately converging to a set of 168 prototypes. This phase serves as a critical refining and pruning of the prototype set, aiming to remove redundancy and form a more compact representation of the ID data, which is essential for establishing a strict OOD decision boundary.

\section{Conclusion}
\label{sec:conclusion}

In this work, we propose PID, a novel dynamic prototypical learning framework for OOD detection. PID aims to address the core limitation of prior work that relies on manually setting a fixed number of prototypes. Our method innovatively introduces a dynamic prototype controller within the MAP-EM framework, which automatically adjusts the number of prototypes for each class according to its intrinsic complexity via birth and death mechanisms. This approach enables PID to learn a more compact and precise representation of the ID data, thereby achieving superior OOD detection performance. Extensive experimental results demonstrate that PID achieves SOTA performance on both Far-OOD and Near-OOD benchmarks. Ablation studies confirm the synergistic necessity of the birth and death mechanisms, while the analysis of prototype count evolution validates the effectiveness of our strategy. A limitation is that the dynamic mechanism introduces two new, relatively sensitive hyperparameters. Future work could focus on investigating adaptive threshold strategies to reduce this sensitivity and further enhance the model's automation.
\FloatBarrier

{
    \small
    \bibliographystyle{ieeenat_fullname}
    \bibliography{main}

@String(NIPS= {Adv. Neural Inform. Process. Syst.})

@String(NIPS  = {NeurIPS})

@article{cui2022out,
  title={Out-of-distribution (ood) detection based on deep learning: A review},
  author={Cui, Peng and Wang, Jinjia},
  journal={Electronics},
  volume={11},
  number={21},
  pages={3500},
  year={2022},
  publisher={MDPI}
}

@article{intro:early-method,
  title={Concrete problems in AI safety},
  author={Amodei, Dario and Olah, Chris and Steinhardt, Jacob and Christiano, Paul and Schulman, John and Mané, Dan},
  journal={arXiv preprint arXiv:1606.06565},
  year={2016}
}

@article{liu2021towards,
  title={Towards out-of-distribution generalization: A survey},
  author={Liu, Jiashuo and Shen, Zheyan and He, Yue and Zhang, Xingxuan and Xu, Renzhe and Yu, Han and Cui, Peng},
  journal={arXiv preprint arXiv:2108.13624},
  year={2021}
}

@article{lu2024learning,
  title={Learning with mixture of prototypes for out-of-distribution detection},
  author={Lu, Haodong and Gong, Dong and Wang, Shuo and Xue, Jason and Yao, Lina and Moore, Kristen},
  journal={arXiv preprint arXiv:2402.02653},
  year={2024}
}

@inproceedings{peng2025distributional,
  title={Distributional Prototype Learning for Out-of-distribution Detection},
  author={Peng, Bo and Lu, Jie and Zhang, Yonggang and Zhang, Guangquan and Fang, Zhen},
  booktitle={Proceedings of the 31st ACM SIGKDD Conference on Knowledge Discovery and Data Mining V. 1},
  pages={1104--1114},
  year={2025}
}

@inproceedings{li2025dpu,
  title={Dpu: Dynamic prototype updating for multimodal out-of-distribution detection},
  author={Li, Shawn and Gong, Huixian and Dong, Hao and Yang, Tiankai and Tu, Zhengzhong and Zhao, Yue},
  booktitle={Proceedings of the Computer Vision and Pattern Recognition Conference},
  pages={10193--10202},
  year={2025}
}

@article{gilmore2000cell,
  title={Cell birth, cell death, cell diversity and DNA breaks: how do they all fit together?},
  author={Gilmore, Edward C and Nowakowski, Richard S and Caviness, Verne S and Herrup, Karl},
  journal={Trends in neurosciences},
  volume={23},
  number={3},
  pages={100--105},
  year={2000},
  publisher={Elsevier}
}

@inproceedings{nguyen2015deep,
  title={Deep neural networks are easily fooled: High confidence predictions for unrecognizable images},
  author={Nguyen, Anh and Yosinski, Jason and Clune, Jeff},
  booktitle={Proceedings of the IEEE conference on computer vision and pattern recognition},
  pages={427--436},
  year={2015}
}

@article{c:25,
  title={A baseline for detecting misclassified and out-of-distribution examples in neural networks},
  author={Hendrycks, Dan and Gimpel, Kevin},
  journal={arXiv preprint arXiv:1610.02136},
  year={2016}
}

@article{c:43,
  title={Enhancing the reliability of out-of-distribution image detection in neural networks},
  author={Liang, Shiyu and Li, Yixuan and Srikant, Rayadurgam},
  journal={arXiv preprint arXiv:1706.02690},
  year={2017}
}

@article{c:37,
  title={Energy-based out-of-distribution detection},
  author={Liu, Weitang and Wang, Xiaoyun and Owens, John and Li, Yixuan},
  journal={Advances in neural information processing systems},
  volume={33},
  pages={21464--21475},
  year={2020}
}

@inproceedings{c:44,
  title={Energy-based open-world uncertainty modeling for confidence calibration},
  author={Wang, Yezhen and Li, Bo and Che, Tong and Zhou, Kaiyang and Liu, Ziwei and Li, Dongsheng},
  booktitle={Proceedings of the IEEE/CVF International Conference on Computer Vision},
  pages={9302--9311},
  year={2021}
}

@article{beyer2020we,
  title={Are we done with imagenet?},
  author={Beyer, Lucas and H{\'e}naff, Olivier J and Kolesnikov, Alexander and Zhai, Xiaohua and Oord, A{\"a}ron van den},
  journal={arXiv preprint arXiv:2006.07159},
  year={2020}
}

@article{c:45,
  title={Your classifier is secretly an energy based model and you should treat it like one},
  author={Grathwohl, Will and Wang, Kuan-Chieh and Jacobsen, J{\"o}rn-Henrik and Duvenaud, David and Norouzi, Mohammad and Swersky, Kevin},
  journal={arXiv preprint arXiv:1912.03263},
  year={2019}
}

@article{c:38,
  title={On the importance of gradients for detecting distributional shifts in the wild},
  author={Huang, Rui and Geng, Andrew and Li, Yixuan},
  journal={Advances in Neural Information Processing Systems},
  volume={34},
  pages={677--689},
  year={2021}
}

@article{c:40,
  title={Csi: Novelty detection via contrastive learning on distributionally shifted instances},
  author={Tack, Jihoon and Mo, Sangwoo and Jeong, Jongheon and Shin, Jinwoo},
  journal={Advances in neural information processing systems},
  volume={33},
  pages={11839--11852},
  year={2020}
}

@article{c:30,
  title={Non-parametric outlier synthesis},
  author={Tao, Leitian and Du, Xuefeng and Zhu, Xiaojin and Li, Yixuan},
  journal={arXiv preprint arXiv:2303.02966},
  year={2023}
}

@article{c:41,
  title={A simple unified framework for detecting out-of-distribution samples and adversarial attacks},
  author={Lee, Kimin and Lee, Kibok and Lee, Honglak and Shin, Jinwoo},
  journal={Advances in neural information processing systems},
  volume={31},
  year={2018}
}

@article{c:42,
  title={Ssd: A unified framework for self-supervised outlier detection},
  author={Sehwag, Vikash and Chiang, Mung and Mittal, Prateek},
  journal={arXiv preprint arXiv:2103.12051},
  year={2021}
}

@article{khosla2020supervised,
  title={Supervised contrastive learning},
  author={Khosla, Prannay and Teterwak, Piotr and Wang, Chen and Sarna, Aaron and Tian, Yonglong and Isola, Phillip and Maschinot, Aaron and Liu, Ce and Krishnan, Dilip},
  journal={Advances in neural information processing systems},
  volume={33},
  pages={18661--18673},
  year={2020}
}

@inproceedings{liu2023good,
  title={Good-d: On unsupervised graph out-of-distribution detection},
  author={Liu, Yixin and Ding, Kaize and Liu, Huan and Pan, Shirui},
  booktitle={Proceedings of the sixteenth ACM international conference on web search and data mining},
  pages={339--347},
  year={2023}
}

@inproceedings{lee2022contrastive,
  title={Contrastive regularization for semi-supervised learning},
  author={Lee, Doyup and Kim, Sungwoong and Kim, Ildoo and Cheon, Yeongjae and Cho, Minsu and Han, Wook-Shin},
  booktitle={Proceedings of the IEEE/CVF conference on computer vision and pattern recognition},
  pages={3911--3920},
  year={2022}
}

@article{zeng2021modeling,
  title={Modeling discriminative representations for out-of-domain detection with supervised contrastive learning},
  author={Zeng, Zhiyuan and He, Keqing and Yan, Yuanmeng and Liu, Zijun and Wu, Yanan and Xu, Hong and Jiang, Huixing and Xu, Weiran},
  journal={arXiv preprint arXiv:2105.14289},
  year={2021}
}

@article{li2020prototypical,
  title={Prototypical contrastive learning of unsupervised representations},
  author={Li, Junnan and Zhou, Pan and Xiong, Caiming and Hoi, Steven CH},
  journal={arXiv preprint arXiv:2005.04966},
  year={2020}
}

@article{banerjee2005clustering,
  title={Clustering on the Unit Hypersphere using von Mises-Fisher Distributions.},
  author={Banerjee, Arindam and Dhillon, Inderjit S and Ghosh, Joydeep and Sra, Suvrit and Ridgeway, Greg},
  journal={Journal of Machine Learning Research},
  volume={6},
  number={9},
  year={2005}
}

@article{parmar2023open,
  title={Open-world machine learning: applications, challenges, and opportunities},
  author={Parmar, Jitendra and Chouhan, Satyendra and Raychoudhury, Vaskar and Rathore, Santosh},
  journal={ACM Computing Surveys},
  volume={55},
  number={10},
  pages={1--37},
  year={2023},
  publisher={ACM New York, NY}
}

@article{c:62,
  title={Enhancing out-of-distribution detection via diversified multi-prototype contrastive learning},
  author={Jia, Yulong and Li, Jiaming and Zhao, Ganlong and Liu, Shuangyin and Sun, Weijun and Lin, Liang and Li, Guanbin},
  journal={Pattern Recognition},
  volume={161},
  pages={111214},
  year={2025},
  publisher={Elsevier}
}

@article{devries2018learning,
  title={Learning confidence for out-of-distribution detection in neural networks},
  author={DeVries, Terrance and Taylor, Graham W},
  journal={arXiv preprint arXiv:1802.04865},
  year={2018}
}

@inproceedings{zhang2021understanding,
  title={Understanding failures in out-of-distribution detection with deep generative models},
  author={Zhang, Lily and Goldstein, Mark and Ranganath, Rajesh},
  booktitle={International Conference on Machine Learning},
  pages={12427--12436},
  year={2021},
  organization={PMLR}
}

@article{ren2021simple,
  title={A simple fix to mahalanobis distance for improving near-ood detection},
  author={Ren, Jie and Fort, Stanislav and Liu, Jeremiah and Roy, Abhijit Guha and Padhy, Shreyas and Lakshminarayanan, Balaji},
  journal={arXiv preprint arXiv:2106.09022},
  year={2021}
}

@inproceedings{sun2022out,
  title={Out-of-distribution detection with deep nearest neighbors},
  author={Sun, Yiyou and Ming, Yifei and Zhu, Xiaojin and Li, Yixuan},
  booktitle={International conference on machine learning},
  pages={20827--20840},
  year={2022},
  organization={PMLR}
}

@inproceedings{yang2025oodd,
  title={OODD: Test-time Out-of-Distribution Detection with Dynamic Dictionary},
  author={Yang, Yifeng and Zhu, Lin and Sun, Zewen and Liu, Hengyu and Gu, Qinying and Ye, Nanyang},
  booktitle={Proceedings of the Computer Vision and Pattern Recognition Conference},
  pages={30630--30639},
  year={2025}
}

@article{xu2025hierarchical,
  title={Hierarchical Multi-Prototype Discrimination: Boosting Support-Query Matching for Few-Shot Segmentation},
  author={Xu, Wenbo and Huang, Huaxi and Gong, Yongshun and Yu, Litao and Wu, Qiang and Zhang, Jian},
  journal={IEEE Transactions on Multimedia},
  year={2025},
  publisher={IEEE}
}

@inproceedings{netzer2011reading,
  title={Reading digits in natural images with unsupervised feature learning},
  author={Netzer, Yuval and Wang, Tao and Coates, Adam and Bissacco, Alessandro and Wu, Baolin and Ng, Andrew Y and others},
  booktitle={NIPS workshop on deep learning and unsupervised feature learning},
  volume={2011},
  number={5},
  pages={7},
  year={2011},
  organization={Granada}
}

@article{yu2015lsun,
  title={Lsun: Construction of a large-scale image dataset using deep learning with humans in the loop},
  author={Yu, Fisher and Seff, Ari and Zhang, Yinda and Song, Shuran and Funkhouser, Thomas and Xiao, Jianxiong},
  journal={arXiv preprint arXiv:1506.03365},
  year={2015}
}

@inproceedings{cimpoi2014describing,
  title={Describing textures in the wild},
  author={Cimpoi, Mircea and Maji, Subhransu and Kokkinos, Iasonas and Mohamed, Sammy and Vedaldi, Andrea},
  booktitle={Proceedings of the IEEE conference on computer vision and pattern recognition},
  pages={3606--3613},
  year={2014}
}

@article{zhou2017places,
  title={Places: A 10 million image database for scene recognition},
  author={Zhou, Bolei and Lapedriza, Agata and Khosla, Aditya and Oliva, Aude and Torralba, Antonio},
  journal={IEEE transactions on pattern analysis and machine intelligence},
  volume={40},
  number={6},
  pages={1452--1464},
  year={2017},
  publisher={IEEE}
}

@article{xu2015turkergaze,
  title={Turkergaze: Crowdsourcing saliency with webcam based eye tracking},
  author={Xu, Pingmei and Ehinger, Krista A and Zhang, Yinda and Finkelstein, Adam and Kulkarni, Sanjeev R and Xiao, Jianxiong},
  journal={arXiv preprint arXiv:1504.06755},
  year={2015}
}

@article{mclachlan1999mahalanobis,
  title={Mahalanobis distance},
  author={McLachlan, Goeffrey J},
  journal={Resonance},
  volume={4},
  number={6},
  pages={20--26},
  year={1999}
}

@inproceedings{wang2022vim,
  title={Vim: Out-of-distribution with virtual-logit matching},
  author={Wang, Haoqi and Li, Zhizhong and Feng, Litong and Zhang, Wayne},
  booktitle={Proceedings of the IEEE/CVF conference on computer vision and pattern recognition},
  pages={4921--4930},
  year={2022}
}

@inproceedings{du2022towards,
  title={Towards unknown-aware learning with virtual outlier synthesis},
  author={Du, Xuefeng and Wang, Zhaoning and Cai, Mu and Li, Sharon},
  booktitle={International Conference on Learning Representations},
  volume={1},
  number={3},
  pages={5},
  year={2022}
}

@article{ming2022exploit,
  title={How to exploit hyperspherical embeddings for out-of-distribution detection?},
  author={Ming, Yifei and Sun, Yiyou and Dia, Ousmane and Li, Yixuan},
  journal={arXiv preprint arXiv:2203.04450},
  year={2022}
}

@techreport{krizhevsky2009learning,
  title={Learning multiple layers of features from tiny images},
  author={Krizhevsky, Alex and Hinton, Geoffrey and others},
  institution={University of Toronto},
  year={2009},
  address={Toronto, ON, Canada}
}
}


\end{document}